%% file: iclr2023_conference_tinypaper.tex
\documentclass{article} % For LaTeX2e
\usepackage{iclr2023_conference_tinypaper,times}
\usepackage{booktabs}
\usepackage{graphicx}
\usepackage{subcaption}
\input{math_commands.tex}

\usepackage{hyperref}
\usepackage{url}

\title{G-PECNet: Towards a Generalizable Pedestrian Trajectory Prediction System}

\author{Aryan Garg \\
\texttt{aryangarg019@gmail.com} \\
\And
Renu M. Rameshan \\  
Lead Research Scientist\\
Vehant Technologies, India\\
\texttt{renur@vehant.com}
}

\iclrfinalcopy % Uncomment for camera-ready version, but NOT for submission.
\begin{document}

\maketitle

\begin{abstract}
Navigating dynamic physical environments without obstructing or hurting humans is of quintessential importance for social robots. In this work, we solve autonomous \textit{drone} navigation's sub-problem of predicting out-of-domain human and agent trajectories using a deep generative model.  
Here, we introduce General-PECNet that improves $9.5\%$ on the Final Displacement Error (FDE) on 2020's benchmark: PECNet~\citep{mainpaper} through a combination of architectural improvements inspired by periodic activation functions~\citep{SIREN} and synthetic trajectory (data) augmentations using Hidden Markov Models (HMMs) and Reinforcement Learning (RL). Additionally, we propose a simple geometry-inspired loss and evaluation metric for trajectory non-linearity analysis. 
% We perform a comprehensive analysis on 
% PECNet was evaluated on a novel geometry-inspired metric and out-of-domain (stress) adapted using synthetic data. Data augmentations . 
% We also propose three improvements for automatic robustness and deployment fitness of learning-based methods. 
\end{abstract}

\section{Introduction}
\label{sec:intro}
Multimodal human or pedestrian trajectory prediction is an ill-posed problem of predicting the final and intermediate steps of some or all pedestrians when only a limited context of their previous trajectories and the scene is known. 
This is further complicated by implicit personal values and social rules that pre-define the pedestrians' interaction. Autonomous navigation for robots or social agents~\citep{social_agents}, can only be enabled by accurate predictions for further downstream planning tasks.
For the prediction problem, we contribute a) a novel reinforcement learning-based synthetic dataset and b) a variational autoencoder~\citep{VAE_Kingma} based pedestrian prediction network, which achieves state-of-the-art performance on the goal-point or final destination prediction error (FDE). G-PECNet is an improved adaptation of PECNet~\citep{mainpaper}.

\let\thefootnote\relax\footnotetext{\textsuperscript{$\dagger$} This work was done at the Indian Institute of Technology, Mandi.  Code:  \href{https://github.com/Aryan-Garg/GPECNet}{Github}}

\section{Method}
\label{method}
% To move the original PECNet~\citep{mainpaper} out of a pareto-optimal solution for SDD~\cite{SDD}, we make the following changes:

\subsection{Augmenting with RL Synthetic Trajectories}
\label{sec:syn_data}
\begin{figure*}
    \begin{center}
    \includegraphics[width=0.3\linewidth]{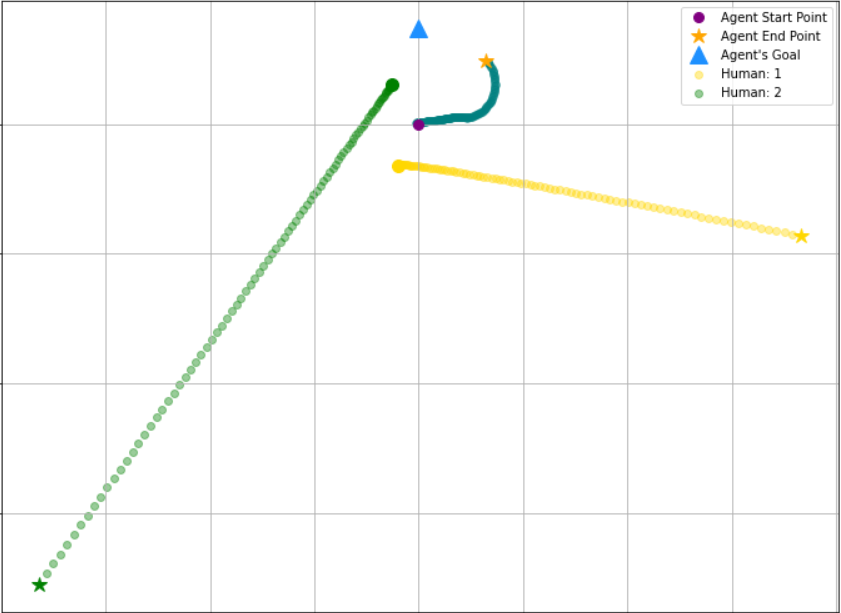}
    \quad
    \includegraphics[width=0.3\linewidth]{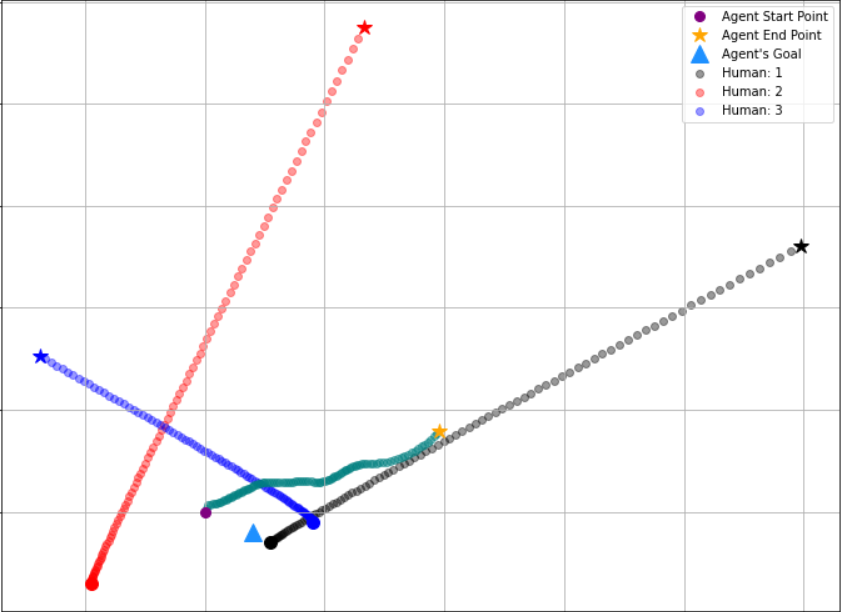} \quad
    \includegraphics[width=0.3\linewidth]{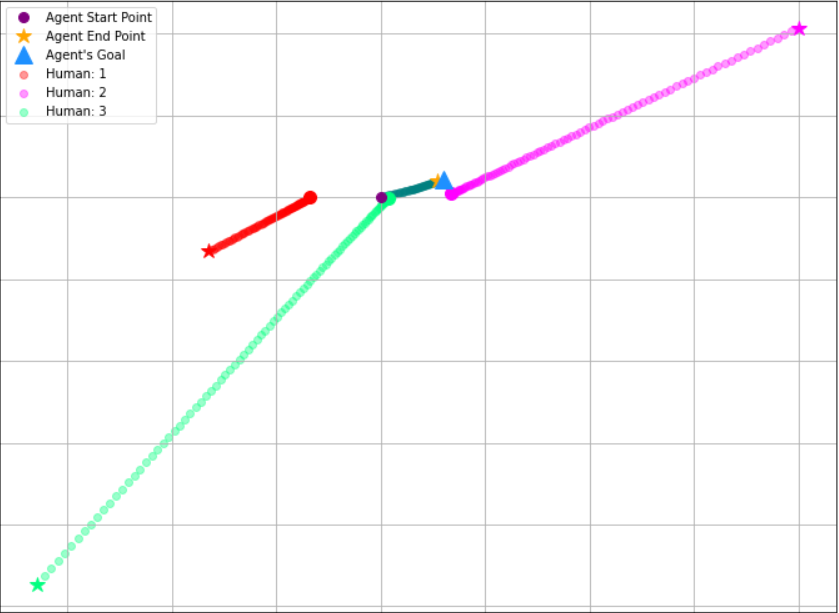}
    \caption{The RL agent is inserted and trained in an HMM interaction playground. Agent's trajectory is turquoise. Evolution of the samples produced. First, the agent learns to turn. The second depicts a complicated scene where the agent learns to avoid multiple collisions. The last scene depicts the agent successfully avoiding a collision and reaching its goal.}
    \label{fig:RL_samples}
    \end{center}
\end{figure*}

Synthetic trajectories were created using traditional Newtonian equations of motion and interaction modeling using a Hidden Markov Model. Finally, we train RL-based bots/agents deployed in the aforementioned interaction (HMM) model using Deep Policy Gradients (DPG). DPG agents were modeled with two major goals: reaching the destination quickly and avoiding collisions with fellow agents/pedestrians. 
% Agents were given control of their acceleration instead of their velocity to simulate smoother, human-like trajectories. 
Apart from acceleration, stopping for another crossing pedestrian (being considerate) was implicitly decided by the agent's \textit{randomly pre-defined} sociability, fitness, and patience attributes. 
% We observe that agents initially learn the greedy approach to reach their goals as fast as possible leading to multiple collisions which are then heavily penalized by the learning objective. 
% This in turn leads to agents choosing sub-optimal paths that avoid maximal collisions. 
We add a circular proximity (fixed radius) detection mechanism to penalize agents that collide with others in the playground. 
% This leads to natural trajectories where agents slow down or speed up when in the proximity of other agents who might potentially collide.
Mathematically, the reward function at time step $t$: $R_t$ for the agents to finally reach the goal $G$ is defined by $
        R_t\: =\: AF^t\, (n_{ICS}+1)^{(AS+AP)} /\, t^2\,(1+\| G - x_t \|_2)
        \label{eq:reward}
$ where $AF$, $AP$, and $AS$ $\epsilon$ $[0,1]$. $x_t$ is the agent's current position, 
and $n_{ICS}$ is the number of impending collision states.
$AS$ and $AP$ are its sociability and patience respectively, determining its recklessness. 
$AF$: Agent's Fitness enforces reaching the goal quickly.
Finally, the loss function is the one used in standard deep-policy gradients methods: $
    J(\phi) = -\,\sum_{t=1}^{t=N}\,log(P_{\phi}(a_t | s_t))\,R_t
$ where, $P(.)$ is parameterized by $\phi$, a simple neural network that emulates the agent's action and state space at any time $t$. Training evolution is shown in figure \ref{fig:RL_samples}.
    
\subsection{Periodic Activation: SIREN Improvement}
\label{sec:siren}

We adapt PECNet to capture finer spatial and temporal details~\citep{SIREN} by replacing all ReLU~\citep{relu_act} activations with a simple sinusoidal function: $\mathbf{x_i} \rightarrow \phi(\mathbf{x_i}) = sin(\mathbf{W_i}\mathbf{x_i} + \mathbf{b_i})$, where $i$ denotes the $i^{th}$ layer of the neural network. This choice is motivated by the finding from ~\cite{SIREN} that current neural network activations are insufficient for modeling high-frequency signals. They fail to represent a signal’s spatial and temporal derivatives which are essential for the solution to the implicitly defined partial differential equations. We notice significant gains when SIRENs are added after our data augmentation. FDE improves by $41.4\%$ when SIREN-infused PECNet is trained with the $6\%$ augmented dataset compared to PECNet on our augmented dataset. See the ablation Tab.~\ref{tab:only_data_aug}. Furthermore, we present a quantitative comparison in~\ref{table:sotas}.

\begin{table}[ht]
\begin{center}
\begin{tabular}{lccc}
\toprule
Learning Rate & ADE & FDE & Best FDE epoch \\
\midrule
0.001 & \textbf{22.20} & 9.32 & 915 \\   
0.0005 & 29.91 &  9.05 & 834  \\   
0.0003 & 25.92 & 9.37 & 998 \\   
0.0002 & 26.75 & \textbf{9.04} & 908  \\   
0.0001 & 25.57 & 9.05 & \textbf{235}  \\  
\bottomrule
\end{tabular}
\end{center}
\caption{State-of-the-art FDE of GPECNet: Trained on 6\% augmented SDD with no standardization and decoupled ADE and FDE for 1000 epochs. We observe that no social pooling results in higher ADE.}
\label{table:no_std}
\end{table}

\subsection{Novel Loss and Evaluation Metric: \textit{AbScore}}
\label{sec:abruptness}
We introduce a simple criterion: Abruptness Score or \textit{AbScore} to measure the turns and variability or non-linearity in each trajectory. An areal-scaled (bounding box area) of the metric is used for outlier detection (data cleaning) and assisted our synthetic dataset creation process. The \textit{AbScore} statistics for SDD trajectories are in Tab.~\ref{table:abscore}. Note that we do not use \textit{AbScore} for training GPECNet. See \ref{sec:app-abrupt} for further mathematical formulation and motivations. 

\begin{table}[ht]
\begin{center}
\begin{tabular}{lc}
\toprule
Statistic & Value \\
\midrule
Max. AbScore & 494866.37 \\  
Min. AbScore & 0.0 \\
Mean & 3430.665 \\  
Std. Deviation & 11987.34 \\  
\bottomrule
\end{tabular}
\end{center}
\caption{Abruptness Score statistics of SDD: A novel loss and evaluation metric for quantifying trajectories' non-linearity.}
\label{table:abscore}
\end{table}
\newpage
\subsection*{URM Statement}
The authors acknowledge that the first author of this work meets the URM criteria of ICLR 2024 Tiny Papers Track.

\bibliography{iclr2023_conference_tinypaper}
\bibliographystyle{iclr2023_conference_tinypaper}

\appendix
\section{Appendix}
The code is available at \href{https://github.com/ANonyMouxe/GPECNet}{Anonymous-Github-repo}. All datasets, created and augmented, will be released upon publication.

\subsection{Quantitative Comparison}
\label{rel}
We present a concise summary of previous seminal pedestrian prediction networks in Tab~\ref{table:sotas}. All previous works use the following 3 datasets: ETH~\citep{ETHDataset}, UCY~\citep{UCYDataset}, and Stanford Drone dataset or SDD~\citep{SDD}.

\begin{table}[h]
\begin{center}
\begin{tabular}{l|cc}
\toprule
Method  & ADE & FDE \\
\midrule
DESIRE~\citep{lee2017desire} & 19.25 & 34.05 \\ 
Social GAN~\citep{gupta2018socialgan}  & 27.23 & 41.44  \\  
Sophie~\citep{sadeghian2018sophie} & 16.27 & 29.38  \\ 
CGNS~\citep{li2019conditional} & 15.6 & 28.2 \\  
CF-VAE~\citep{cfvae} & 12.60 &	22.30  \\  
P2TIRL~\citep{deo2021trajectory} & 12.58 & 22.07 \\  
PECNet~\citep{mainpaper} & 9.96 &	15.88 \\ 
Y-Net~\citep{mangalam2020goals}  & 7.85 & 11.85 \\  
$V^2$-Net~\citep{wong2022view}  & 7.12 &	11.39 \\  
NSP-SFM~\citep{yue2023human} & \textbf{6.52} & 10.61 \\ 
\midrule
\textbf{G-PECNet} & 26.75 & \textbf{9.04} \\
\bottomrule
\end{tabular}
\end{center}
\caption{ADE is the average displacement error and FDE is the final displacement error. 
All networks are evaluated on original SDD~\citep{SDD}, with the total number of pedestrians to consider for predictions as 20 ($K$ = 20), except ours.  
Our ADE ($26.75$) is not low as we use a decoupled PECNet; not using the social pooling layers~\citep{Alahi_2016_CVPR_metrics}.
We primarily focus on predicting the goal point of pedestrians. 
The intuition is that all intermediate steps could then be refined from coarser estimates after the endpoint is fixed, similar to the training procedure of denoising diffusion probabilistic models~\citep{ddpm}.}
\label{table:sotas}
\end{table}

\subsection{Stanford Drone Dataset: Data Analysis}
\label{data_analysis}
Based on the unique quantitative (table: \ref{table:unique_points}), we augmented the training dataset to keep the statistical properties of the training dataset intact.
We perform a classification of the training dataset based on the number of unique points in each trajectory. See table \ref{table:unique_points}.

Then we manually identify 7 qualitative classes for each trajectory as follows:
 \begin{itemize}
 \label{list:classification}
     \item Type 1: Stationary pedestrians
     \item Type 2: 3-8 unique points in trajectory bounded in a 5x5 box 
     \item Type 2F: F means category Flying: A straight line trajectory will be shifted into the line starting from a new point if the perspective of the viewer (here: drone) changes. We call this scenario the flying category trajectory. The drone usually translates along an axis here.
     \item Type 3: 3-9 unique points loosely bounded in a 100x100 box
     \item Type 3F: (Flying) Same as 2F
     \item Type 4: Start and Goal points are within a 5x5 box
     \item Type 5: Flying randomly. This is different from 2F and 3F in the sense that the drone translated haphazardly here.
     \item Type 6: Backtracker: The pedestrian re-traces his steps after a while. Usually after 6-7 steps.
     \item Type 7: Perfectly Linear to Moderately linear trajectories that could be modeled by simple Newtonian mechanics.
 \end{itemize}
 
These two classifications were done to emulate the statistical properties of the training dataset for augmentation purposes. Based on this data analysis and classifications (Tab.~\ref{table:unique_points} and Sec.~\ref{data_analysis} respectively), we augmented the training dataset using a Deep Policy Gradient Network~\ref{dpg} + HMM interaction model and some Newtonian trajectories to keep the statistical properties of the training dataset intact. 
We manually discard any trajectories that did not fit the 7 types of trajectories identified in~\ref{list:classification}.

\subsection{Deep Policy Gradient Network}
\label{dpg}
\begin{figure}[ht]
    \centering
    \includegraphics[width=0.7\linewidth]{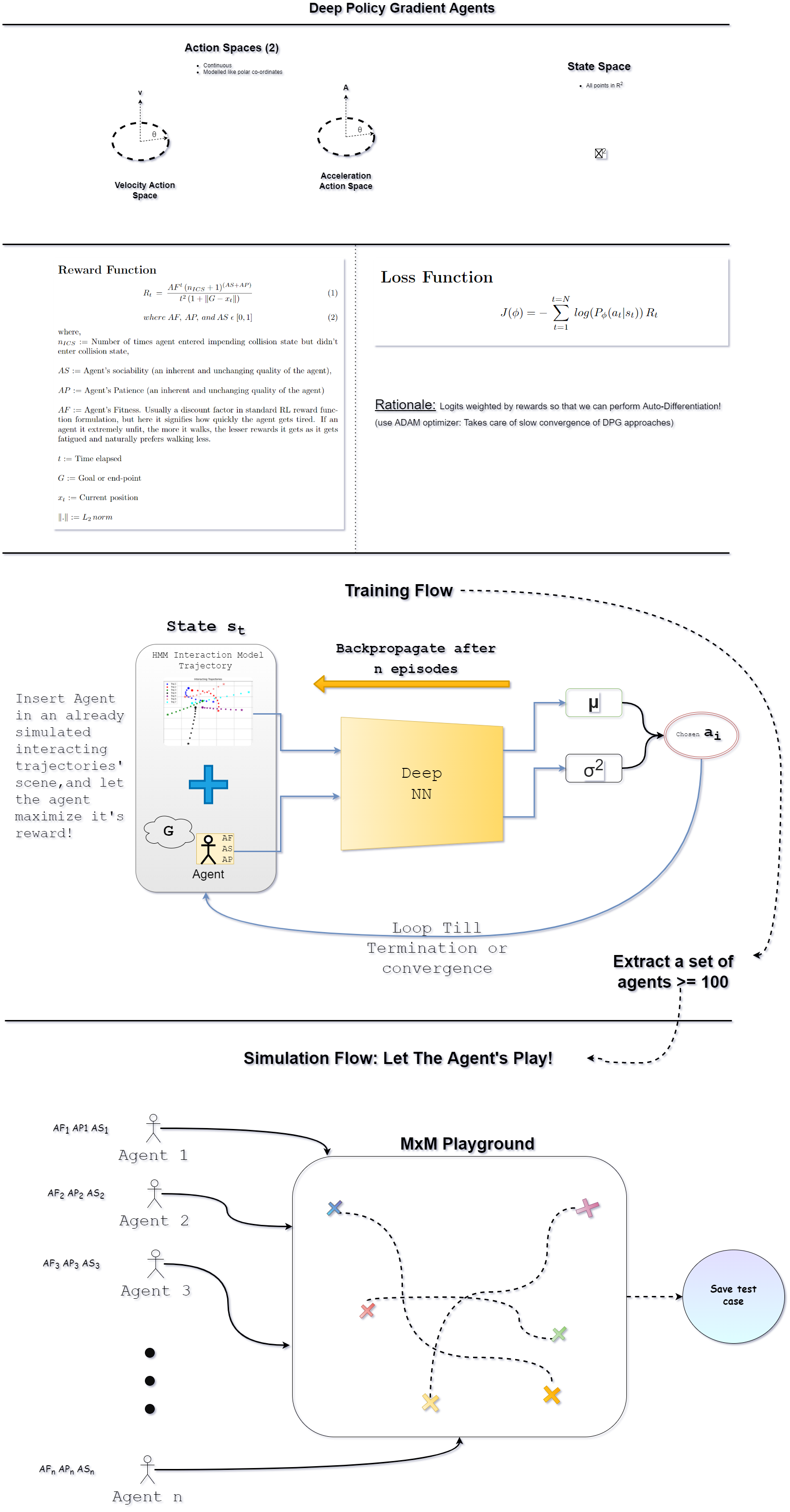}
    \caption{RL modeling}
    \label{fig:enter-label}
\end{figure}

We use a simple fully connected ReLU-activated neural network (nodes: 8 $\rightarrow$ 16 $\rightarrow$ 8 $\rightarrow$ 4) with 8 inputs: current $x$-coordinate, current $y$-coorinate, $x$-goal, $y$-goal, fitness, patience, sociability, and distance to nearest person/agent and 4 output nodes defining the action space: the speed, direction, acceleration magnitude and acceleration direction to take another step. An overview of the whole workflow can be found in Fig~\ref{fig:enter-label}.

HMMs were considered for the interaction modeling due to their high success in spatiotemporal tasks~\citep{HMMs}.

\begin{table}[ht]
\begin{center}
\begin{tabular}{|l|c|c|}
\hline
Unique Points & Trajectories & \% dataset \\
\hline\hline
1 & 145 & 5.13\% \\  
2 & 62 & 2.19 \% \\  
3 & 71 & 2.51 \% \\  
4 & 69& 2.44\% \\  
5 & 57& 2.01\% \\  
6 & 41& 1.45\% \\  
7 & 51& 1.80\%  \\  
8 & 28& 0.99\% \\  
9 & 26 & 0.92\%  \\  
10 & 24 & 0.85\%  \\  
11 & 22 & 0.78\%  \\  
12 & 25 & 0.88\%  \\  
13 & 17 & 0.60\%  \\  
14 & 24 & 0.85\%  \\  
15 & 22 & 0.78\%  \\  
16 & 22 & 0.78\%  \\  
17 & 39 & 1.38\%  \\  
18 & 30 & 1.06\%  \\  
19 & 76 & 2.69\%  \\  
20 & 1978 & 69.92\% \\  
\hline
\end{tabular}
\end{center}
\caption{SDD: Trajectories' unique points. 145 trajectories had 1 unique point, i.e, the goal and starting point as the same with all other points being sampled there itself: Stationary}
\label{table:unique_points}
\end{table}

\subsection{Architecture Brief}
PECNet integrates a Conditional Variational Autoencoder (CVAE) while infusing probabilistic elements into the trajectory generation process. PECNet is equipped with specialized components, including 3 dedicated encoders for a) the past trajectories, b) the destination c) a latent space encoder, and finally a predictor for forecasting future trajectories. We depart from PECNet by using custom sinusoidally activated multi-layer perceptrons (MLPs). The incorporation of sine activations is a departure from traditional ReLU and variants to capture high frequency spatial and temporal details of a signal, as also demonstrated in~\citep{SIREN}. Notably, PECNet employs non-local social pooling mechanisms facilitated by three critical MLPs named: non-local-theta, non-local-phi, and non-local-g. They capture intricate long-range interactions among pedestrians. Since we do not use these networks or decouple the system, we see a high average displacement error. 

During training, the model utilizes destination information to produce diverse and probabilistic future trajectories. During inference, it predicts future trajectories given historical context only. Please refer to our \href{https://github.com/ANonyMouxe/GPECNet/blob/main/GPECNet/utils/models_SIREN.py}{codebase} for exact parameters and layer definitions. 

\subsection{Further Experiments}

\subsubsection{Ablation: Decoupled ADE \& FDE or No Social Pooling in PECNet}

We performed two ablation studies. First, by decoupling the ADE and the FDE metrics. See table \ref{table:decoupled}.

\begin{table}[ht]
\begin{center}
\begin{tabular}{|l|c|c|c|}
\hline
Learning Rate & ADE & FDE & Best FDE epoch \\
\hline\hline
0.001 & $>$50 & 15.68 & 457 \\   
0.0005 & $>$50 & 15.76 & 301  \\   
0.0003 & $>$50 &  15.9 & 541 \\   
0.0002 & $>$50 & 15.65 & 420 \\   
0.0001 & $>$50 & 15.92 & 391  \\  
\hline
\end{tabular}
\end{center}
\caption{Decoupled or without social pooling - PECNet (ADE and FDE) trained on original SDD with different learning rates. Here, ADE is independent of FDE. No SIREN improvement or data augmentations were applied either.}
\label{table:decoupled}
\end{table}

% \subsubsection{Ablation: No Standardization}
% Without the standardization parameter on the input trajectories. We note an adverse effect on ADE and a significant state-of-the-art improvement on FDE, by removing standardization. See table \ref{table:no_std}.

% We notice an absurdity in PECNet's codebase where the standardization parameter($=1.86$) is also dividing the ADE metric. We remove that division for G-PECNet and report fair metrics.
\subsubsection{Effects of Data Augmentations}
We sample the RL and Newtonian trajectories in a fashion to keep the statistics in Tab.\ref{table:unique_points} similar. We report the ADE and FDE metrics of various levels of augmentations from $1\%$ to $18\%$ in table \ref{tab:only_data_aug}. 

\begin{figure}
    \centering
    \includegraphics[width=0.8\linewidth]{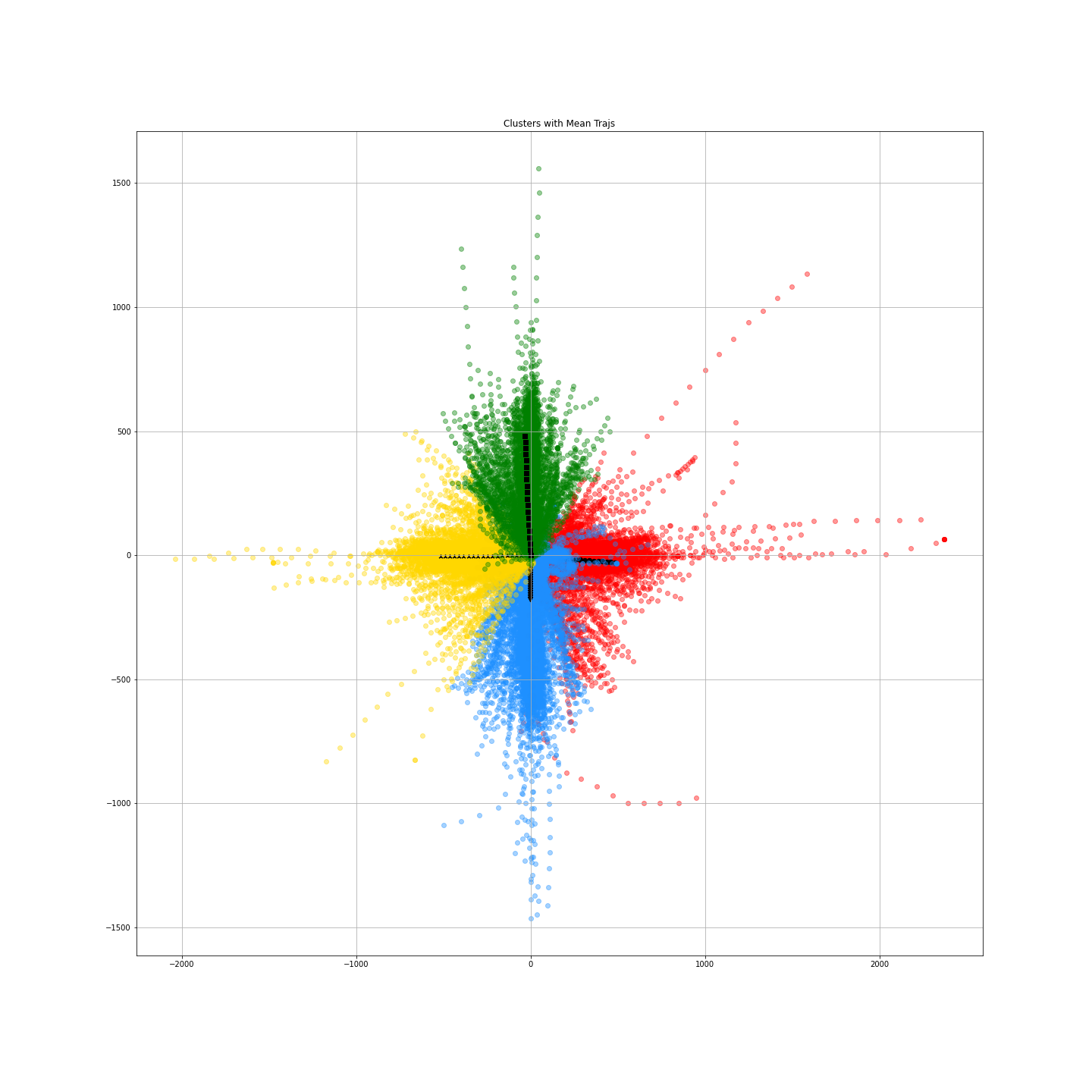}
    \caption{SDD training dataset. Frobenius norm-based clustering of trajectories, with black trajectories representing the cluster.}
    \label{fig:all_SDD}
\end{figure}

We also observe that PECNet heavily overfits on the SDD dataset and it is probable that Adam \cite{kingma2017adam} finds a deep crevice in the gradient surface. An evolutionary optimization strategy like Covariance Matrix Adaptation (CMA-ES) \cite{cma_es} would highlight the shortcomings of the robustness of PECNet. 

\begin{table}[h]
\begin{center}
\begin{tabular}{l|c|c|c}
\toprule
Augment \% & Total Trajectories & ADE & FDE \\
\midrule
1\%   & 18328     & 64.34 & 19.18 \\
3\%   & 19048     & 53.22 & 15.63 \\  
5\%   & 19766     & 45.17 & 15.72 \\  
6\%   & 20126     & 51.73 & \textbf{15.43} \\  
8\%   & 20844     & 46.37 & 15.75 \\  
10\%  & 21564     & 51.51 & 15.54 \\  
13\%  & 22642     & \textbf{40.76} & 15.90 \\  
15\%  & 23360     & 51.40 & 18.36 \\  
18\%  & 24438     & 56.56 & 15.90 \\  
\bottomrule
\end{tabular}
\end{center}
\caption{Effects of augmenting SDD with our synthetic trajectories in varying proportions and training PECNet with it. Standard learning rate: $3e-4$, 1000 epochs, and no social pooling were used across runs. Note that the training dataset has $\sim$18k trajectories due to simple augmentations(rotations \& translations) also saved. This was introduced by \cite{mainpaper}. Since the agent simulations and Newtonian trajectory simulators are inherently random, we decided to sample $18k$ trajectories at once and subsequently used that purely synthetic dataset to sample \& augment SDD. The sampling is deterministic as we always select the first $k\%$ from the fixed ordering. 
}
\label{tab:only_data_aug}
\end{table}

\subsubsection{Baseline Metrics' Reproduction (PECNet)}

\begin{table}[ht]
\begin{center}
\begin{tabular}{|l|c|c|}
\hline
Learning Rate & Best ADE & Best FDE \\
\hline\hline
0.001 & 11.01 & 15.62 \\ 
0.0005 & 12.52 & 15.68 \\  
\textbf{0.0003} & \textbf{10.47} & \textbf{15.60}\\ 
0.0002 & 10.65 & 15.78\\  
0.0001 & 10.65 & 15.78 \\  
\hline
\end{tabular}
\end{center}
\caption{Sanity Run on different learning rates on original Stanford Drone Dataset, with social pooling. Bold represents benchmark reproduction.}
\label{table:lr_sanity}
\end{table}

\subsection{Non-linearity analysis: Abruptness-Score}
\label{sec:app-abrupt}
\begin{figure}
    \centering
    \includegraphics[scale=0.38]{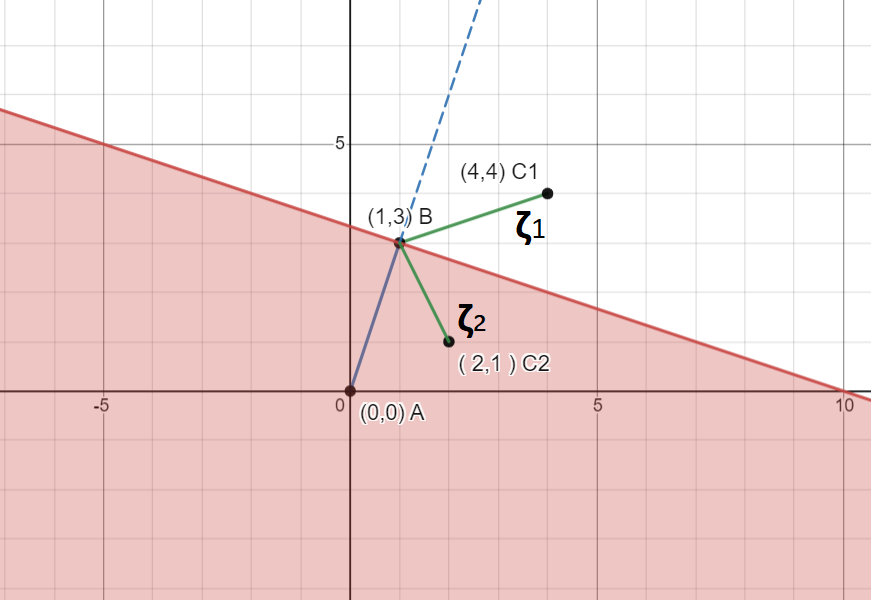}
    \caption{Defining Turns}
    \label{fig:defining_turns}
\end{figure}

We clustered SDD's trajectories (\ref{fig:all_SDD}) based on the bounding boxes to get an estimate of the maximal displacement and turn in each trajectory. 
Based on this information, we introduce a novel and simple metric: \textit{Abruptness Score} to measure the turns and variability or non-linearity in each trajectory. An areal-scaled and an unscaled version of the metric is used for analysis and outlier detection. The intuition and mathematical formulation are as follows: 

In the example figure \ref{fig:defining_turns}, the trajectories $\zeta_1 = \{A, B, C1\}$ and $\zeta_2 = \{A, B, C2\}$ are shown. The dotted blue line is normal to the red danger zone. Points that fall under this danger zone will form an obtuse turn trajectory like $\zeta_2$. Naturally, we want the score to assign a larger value to $\zeta_2$ than $\zeta_1$ since the turn is huge and the trajectory (more) abruptly changes direction. 

Mathematically, 

\begin{equation}
    AbScore = \left \lceil \frac{180.\theta}{10\pi} \right \rceil | \vec{a} \times \vec{b} |
\end{equation}

where 
\begin{equation}
    \vec{a} = \vec{AB} , \vec{b} = \vec{BC}
\end{equation}

\begin{equation}
    \theta = | \arcsin \frac{\vec{a} \times \vec{b}}{|\vec{a}||\vec{b}|}|
\end{equation}

If $\theta$ is obtuse, we add $pi/2$ to $\theta$ before sending it to equation 1.

For scaling we simply divide the abruptness score by the area of the tightest-bounding-box of the trajectory or divide by $(max(\zeta_x)-min(\zeta_x)) * (max(\zeta_y)-min(\zeta_y)$. For perfectly linear trajectories, we use the length of the trajectory for scaling. 

\begin{figure}[h]
    \centering
    \includegraphics[scale=0.5]{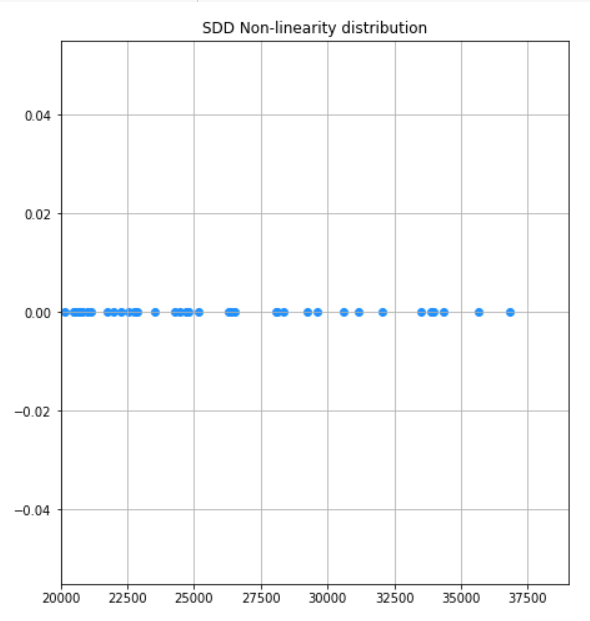}
    \caption{SDD's Non-Linearity Distribution Tail. (Thresholded to remove outliers)}
    \label{fig:abscore_nld}
\end{figure}

We need areal-scaling to get an unbiased estimate of trajectories' non-linearity that spans widely different sizes or regions. 
Based on this metric we analyze SDD and report that the trajectories are not non-linear on average however the distribution contains outliers with huge non-linearity scores. This analysis provided us with an estimate of the dataset's non-linearity for synthetic dataset generation purposes. See table \ref{table:abscore} and figure \ref{fig:abscore_nld}.

\subsection{Discussion}
\label{sec:Discussion}
We demonstrated state-of-the-art final displacement errors on the Stanford Drone Dataset with our method GPECNet. The core improvements originate from our rich synthetic data augmentations coupled with SIRENs~\citep{SIREN} that can capture better high-frequency spatial and temporal dependencies.
Even though our method achieves the best FDE results, the critical nature of systems that could employ our algorithm necessitates introducing a confidence metric for larger controllability and explainability. Another avenue to extend our work is generating multi-modal predictions simultaneously to move towards a real deployable system.

\end{document}

%% file: math_commands.tex
\usepackage{amsmath,amsfonts,bm}

\def\eqref#1{equation~\ref{#1}}
\def\1{\bm{1}}

\DeclareMathAlphabet{\mathsfit}{\encodingdefault}{\sfdefault}{m}{sl}
\SetMathAlphabet{\mathsfit}{bold}{\encodingdefault}{\sfdefault}{bx}{n}